\documentclass[10pt,twocolumn,letterpaper]{article}

\usepackage{cvpr}              % To produce the CAMERA-READY version
\usepackage{graphicx}
\usepackage{amsmath}
\usepackage{amssymb}
\usepackage{booktabs}
\usepackage{soul}
\usepackage{multirow}
\usepackage{amsthm}
\newtheorem*{remark}{Remark}

\usepackage[accsupp]{axessibility} % Improves PDF readability for those with disabilities.

\usepackage{float}
\usepackage[pagebackref,breaklinks,colorlinks]{hyperref}

\usepackage[capitalize]{cleveref}
\crefname{section}{Sec.}{Secs.}
\Crefname{section}{Section}{Sections}
\Crefname{table}{Table}{Tables}
\crefname{table}{Tab.}{Tabs.}

\def\cvprPaperID{5265} % *** Enter the CVPR Paper ID here
\def\confName{CVPR}
\def\confYear{2022}

\begin{document}

\title{ManiTrans: Entity-Level Text-Guided Image Manipulation via Token-wise Semantic Alignment and Generation}

\author{Jianan Wang\textsuperscript{1}
\quad
Guansong Lu\textsuperscript{2}
\quad 
Hang Xu\textsuperscript{2}
\quad
Zhenguo Li\textsuperscript{2}
\quad
Chunjing Xu\textsuperscript{2}
\quad
Yanwei Fu\textsuperscript{1} 
\\
\textsuperscript{1}School of Data Science, Fudan University
\quad
\textsuperscript{2}Huawei Noah’s Ark Lab
\quad
\\
{\tt\small \{jawang19, yanweifu\}@fudan.edu.cn}
\quad
{\tt\small \{luguansong, xu.hang, li.zhenguo, xuchunjing\}@huawei.com}
}

\twocolumn[{%`
\renewcommand\twocolumn[1][]{#1}%
\maketitle
\begin{center}
\centering
\includegraphics[width=\textwidth]{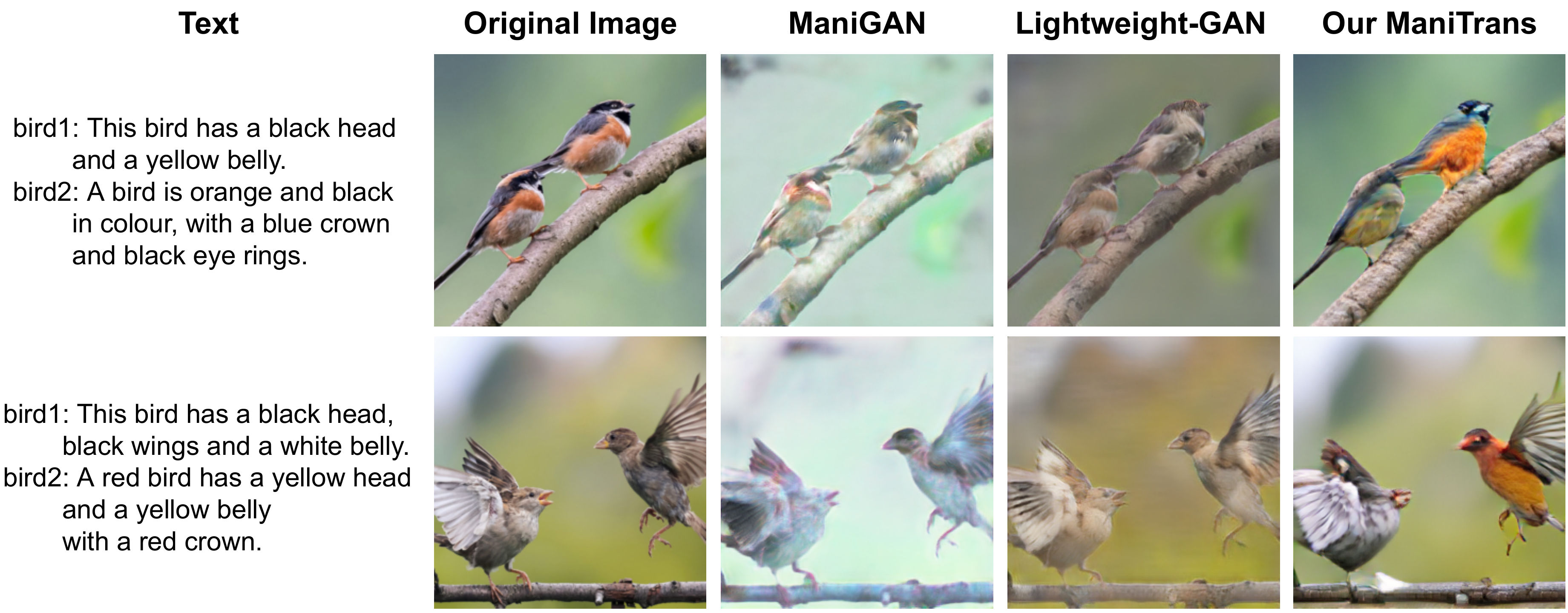}
% \vspace{-0.1in}
\captionof{figure}{
Results of manipulating multiple entities according to different texts with ManiGAN \cite{2019ManiGAN}, Lightweight-GAN \cite{2020Lightweight} and our ManiTrans. Our ManiTrans can manipulate different entities accordingly, while two baseline methods fail.
}
\label{fig:multiple_object}
\end{center}
}]

\begin{abstract}

Existing text-guided image manipulation methods aim to modify the appearance of the image or to edit a few objects in a virtual or simple scenario, which is far from practical application. In this work, we study a novel task on text-guided image manipulation on the entity level in the real world. The task imposes three basic requirements, (1) to edit the entity consistent with the text descriptions, (2) to preserve the text-irrelevant regions, and (3) to merge the manipulated entity into the image naturally. To this end, we propose a new transformer-based framework based on the two-stage image synthesis method, namely \textbf{ManiTrans}, which can not only edit the appearance of entities but also generate new entities corresponding to the text guidance. Our framework incorporates a semantic alignment module to locate the image regions to be manipulated, and a semantic loss to help align the relationship between the vision and language. We conduct extensive experiments on the real datasets, CUB, Oxford, and COCO datasets to verify that our method can distinguish the relevant and irrelevant regions and achieve more precise and flexible manipulation compared with baseline methods. The project homepage is \url{https://jawang19.github.io/manitrans}.

\end{abstract}

\begin{figure*}
\centering
\includegraphics[width=\textwidth]{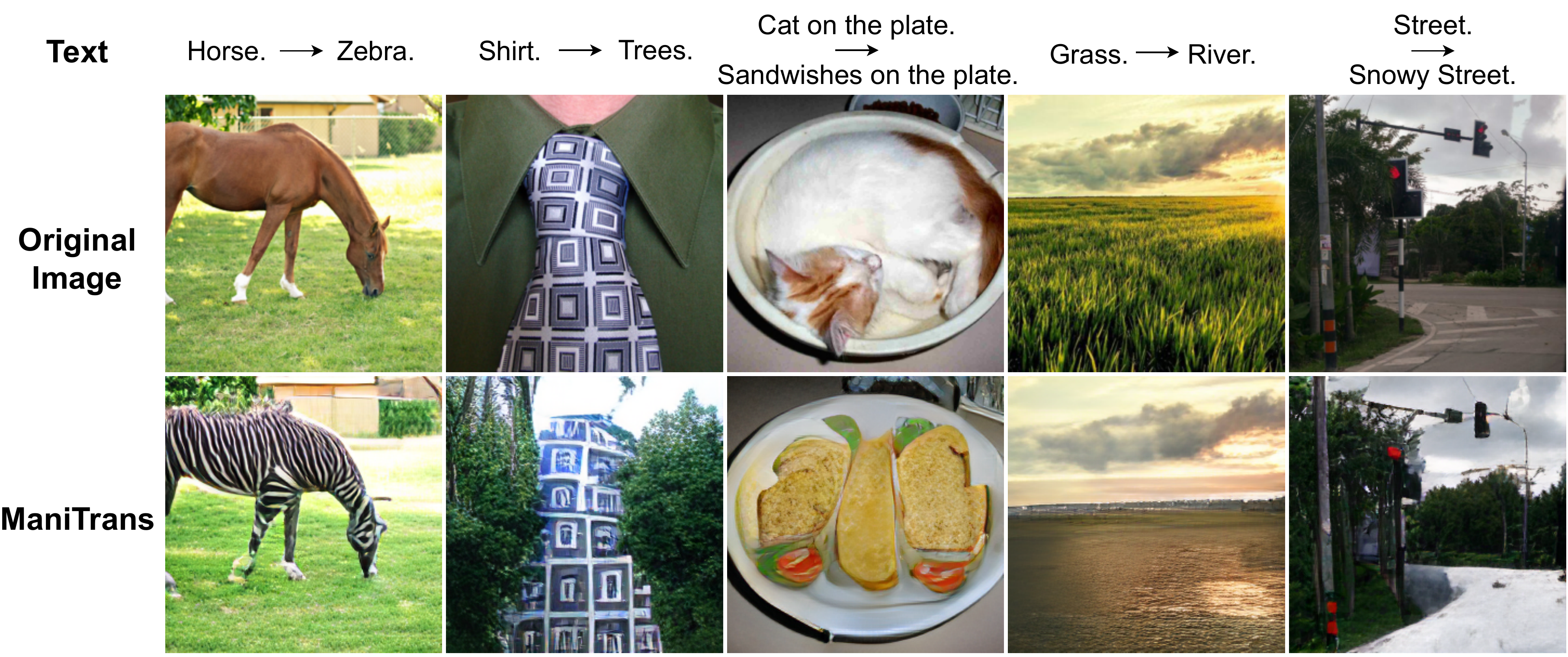}
% \vspace{-0.2in}
\caption{
Results of manipulation results across categories on COCO. The texts follow the expression rule, ``A $\rightarrow$ B'', which indicates that our ManiTrans transfers entity A into another kind of entity B.
}
\label{fig:coco_across_category}
% \vspace{-0.15in}
\end{figure*}

\section{Introduction}

There are various active branches of image manipulation, such as style transfer \cite{style_transfer}, image translation \cite{isola2017pix2pix,zhu2017cyclegan}, 
and Text-Guided Image Manipulation (TGIM),
by taking advantage of recent deep generative architectures such as GANs~\cite{2014Generative}, VAE~\cite{2014Auto} and auto-regressive models~\cite{vaswani2017attention}. 
Particularly, the previous TGIM methods either operate some objects by  text instructions \cite{el2019tell,zhang2021text,fu2020iterative}, such as ``adding'' and ``removing'' in a simple toy scene, or manipulating the appearance 
of objects\cite{chen2018language} or the style of the image\cite{wang2018learning,jiang2021language}.
In this work, we are interested in a novel challenging  task of 
entity-Level Text-Guided Image Manipulation (eL-TGIM), which is to manipulate the entities on a natural image given the text descriptions,
as shown in Fig.~\ref{fig:multiple_object} and Fig.~\ref{fig:coco_across_category}.  Critically, 
our eL-TGIM is much more difficult than the vanilla TGIM task, as it demands much manipulation ability in the fine-grained entity level. Thus, it is nontrivial to directly extend previous methods to this eL-TGIM task, as they can not effectively identify and edit the properties of entities as empirically shown in Fig.~\ref{fig:multiple_object}.

Generally, the major obstacle of the TGIM task lies in distinguishing which parts of the image to change or not change. 
To tackle this problem, existing TGIM methods  \cite{sisgan,tagan,2019ManiGAN,2020Lightweight} 
propose many different manipulation mechanisms, such as word-level discriminator \cite{tagan,2020Lightweight} and text-image affine combination module \cite{2019ManiGAN}, to differentiate the candidate editing regions from the other image parts.
These methods unfortunately are still very limited to be applied to manipulate the entities in nature images. For example,
Fig.~\ref{fig:multiple_object} shows that previous methods can only manipulate the texture/color of an object, while they fail to generate reasonable entity-level manipulation results from the text description.

To this end, we propose a novel framework of Manipulating Transformers (ManiTrans) with the token-wise semantic alignment and generation for the eL-TGIM. Thus, to tackle this task, we propose the two key ideas of \textit{Transformer based image synthesizer (Trans)}, and \textit{entity-level semantic manipulator (Mani)}. 
Specifically, the recent transformer-based
architecture \cite{2017Neural,vqvae2,vqgan} has been proposed for image synthesis and has shown great expressive power. %In the first stage, 
We thus present a novel component of Trans, by first learning an autoencoder to downsample and quantize an image as a sequence of discrete image tokens and then fit the joint distribution of this sequence with a transformer-based auto-regressive model.

Furthermore, to successfully identify the entities for editing, we propose the Mani component which includes a semantic alignment module, and Contrastive Language-Image Pre-training (CLIP) module. The former module helps the generation model Trans to
locate and modify the text-relevant image tokens
given the textual guidance.  Thus our ManiTrans generation model can manipulate the image locally and preserve the irrelevant contents to a greater extent, as in Fig.~\ref{fig:multiple_object}. On the other hand, we repurpose the recent CLIP module as one type of semantic loss to further boost the visual-semantic alignment between the input textual guidance and the manipulated image. Essentially, such semantic loss, proposed in our ManiTrans, is complementary to token-wise classification loss, and thus efficiently serves as a pixel-level supervision signal to train our model. 

We evaluate our method on multiple datasets including: CUB \cite{cub200}, Oxford \cite{oxford102} and COCO \cite{mscoco}. Quantitatively and qualitatively comparison against previous methods shows that our method can better manipulate the entities of an image by text while keeping the background region unchanged. Besides manipulating the texture/color of one object, our method also shows superior capability for manipulating the structure of objects guided by various textual descriptions, as shown in Fig.~\ref{fig:multiple_object} and Fig.~\ref{fig:birdflowerexchange} respectively. This can not be done in previous methods.

In summary, our contributions are as follows:

\noindent $\bullet$ We propose a %two-stage 
transformer-based entity-level
text-guided image manipulation framework with token-wise semantic alignment and generation, named ManiTrans, which can not only manipulate the texture/color of a single object, but also manipulate the structure of an object and manipulate multiple objects. \\
\noindent $\bullet$ We propose a semantic alignment module to locate the text-relevant image tokens for flexible manipulation, and a semantic loss for better visual-semantic alignment and detailed training signal.
\\
\noindent  $\bullet$ We repurpose and utilize the transformer-based image synthesizer, and CLIP module as the semantic loss in our ManiTrans framework, which is nontrivial technically. \\
\noindent $\bullet$ We quantitatively and qualitatively evaluate our method on the CUB, Oxford and COCO datasets, achieving superior/competitive results against baseline methods.

\section{Related work}

\noindent \textbf{Text-to-image Generation}
Text-to-image generation focuses on generating images to visualize what texts describe. There are many good GAN-based models  \cite{GAN-INT-CLS, xu2018attngan,zhang2017stackgan,zhang2018stackgan++}.
Li \etal \cite{li2019controllable} further introduce a word-level discriminator network to provide the generator network with fine-grained feedback. Besides GANs, recent works also explore applying transformer-based network for text-to-image generation \cite{dalle,ding2021cogview,esser2021imagebart}. 
In contrast, rather than generating images according to texts, we focus on entity-level manipulating input images according to texts.

\noindent \textbf{Text-guided Image Manipulation}
Text-guided image manipulation has attracted extensive attention as it enables the users to flexibly edit an image with natural language~\cite{sisgan,tagan,
2019ManiGAN,2020Lightweight,xia2021tedigan,stylegan,el2019tell,zhang2021text,fu2020iterative,chen2018language,wang2018learning,jiang2021language}. 
Particularly,
Li~\etal~\cite{2019ManiGAN} introduces a multi-stage network with a novel text-image combination module to generate high-quality images. Li~\etal~\cite{2020Lightweight} propose a new word-level discriminator along with explicit word-level supervisory labels to provide the generator with detailed training feedback related to each word, achieving a lightweight and efficient generator network.
Recently, due to the good synthesizing capability of StyleGAN, researchers devote to image manipulation by pre-trained StyleGAN models~\cite{xia2021tedigan,patashnik2021styleclip}. Patashnik \etal \cite{patashnik2021styleclip} adopt the CLIP model for semantic alignment between text and image, and propose mapping the text prompts to input-agnostic directions in StyleGAN’s style space, achieving interactive text-driven image manipulation.
On the contrary,  our module for image synthesis is trained from the scratch, rather than built upon pre-trained StyleGAN models. Thus our framework should in principle be much more flexible to be deployed to real-world visual applications.

\noindent \textbf{Semantic Image Synthesis}
The task of semantic image synthesis aims to generate a photo-realistic image from a semantic label. 
Isola \etal \cite{isola2017pix2pix} propose a unify framework based on conditional GANs \cite{mirza2014conditional} for various image-to-image translation tasks, including $Semantic ~ labels \leftrightarrow photo$, $Edges \rightarrow Photo$, $Day \rightarrow Night$, and so on. 
Chen and Koltun \cite{chen2017cascaded_refinement} adopt a modified perceptual loss to synthesize high-resolution images to tackle the instability of adversarial training.
Wang \etal \cite{wang2018pix2pixhd} propose a novel adversarial loss and a new multi-scale generator and discriminator architectures for generating high-resolution images with fine details and realistic textures. Park \etal \cite{park2019SPADE} propose a spatially-adaptive normalization layer to modulate the activations using input semantic layouts and effectively propagate the semantic information throughout the network.
Such works enable users to synthesis images with a finite number of semantic concepts associated with the semantic labels, while our method focuses on manipulating the input images according to the input texts, which is more flexible and with an unlimited number of semantic concepts.

\noindent \textbf{Vision and Language Representation Learning}
Numbers of vision-language pre-training models \cite{radford2021clip,jia2021align,zhang2021vinvl,zhuge2021kaleido,li2020unimo,chen2020uniter,lin2021m6,li2020oscar,su2019vlBERT} learn cross-modal representations for various down-stream tasks, including image-text retrieval, image captioning, visual grounding, and so on. They adopt the network architecture of ResNets \cite{resnet} and/or Transformers \cite{vaswani2017attention,dosovitskiy2020vit}, and mainly use two kinds of learning tasks for pre-training: cross-modal contrastive learning and masked language modeling. Specifically, the recently CLIP \cite{radford2021clip} model is trained on a large-scale dataset, and shows superior performance on zero-shot tasks. We repurpose the CLIP model as one supervision loss to help train our framework for eL-TGIM.

\section{Method}

\begin{figure*}
\centering
\includegraphics[width=\linewidth]{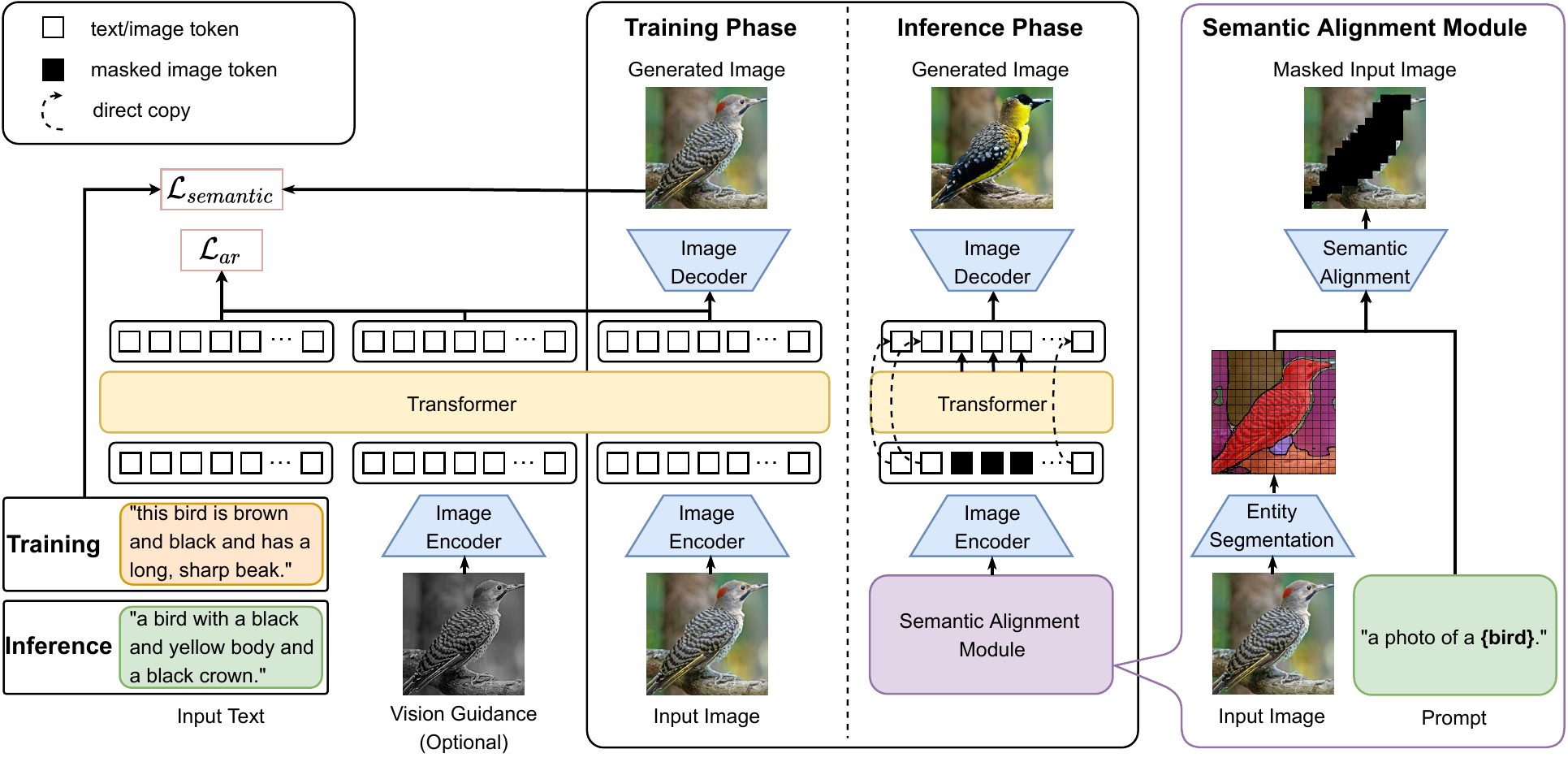}
% \vspace{-0.15in}
\caption{The architecture of our ManiTrans framework for entity-level text-guided image manipulation. The basic idea of ManiTrans is to manipulate only the image tokens which correspond to the text. ManiTrans adopts Transformer as the generation model, which takes the input text tokens and vision guidance tokens (gray/sketch image generated from the input image) as input, and generates image tokens auto-regressively.
Besides the classification loss on each token, ManiTrans adopts a semantic loss to help the model capture the visual-semantic alignment between the input text and the manipulated image in the training phase. In the inference phase, ManiTrans adopts a semantic alignment module to locate the text-relevant tokens to be manipulated and the generation model manipulates only these image tokens. 
The vision guidance is optional and is necessary in the case of editing only the appearance of an entity.
}
\label{fig:framework}
% \vspace{-0.15in}
\end{figure*}

Fig.~\ref{fig:framework} shows the architecture of our ManiTrans framework, which is composed of Mani and Trans. 
In this section, we first introduce the architecture of our model. Then we introduce the language guidance and vision guidance mechanism for model training. Finally, we introduce the semantic alignment module for flexible image manipulation during the inference phase.

\subsection{Manipulating Transformers Model}

Our transformer-based image manipulation model consists of an autoencoder model for downsampling and quantizing the input image as discrete tokens, and a transformer model for fitting the joint distribution of image tokens.

The autoencoder based Trans model consists of three components, a convolutional encoder $E$, a convolutional decoder $G$ and a codebook $\mathbf{Z} \in \mathbb{R}^{K \times n_z}$, 
containing $K$ $n_z$-dimensional latent variables. All of them are learnable.
Given an image $\mathbf{X} \in \mathbb{R}^{H \times W \times 3}$, $E$ encodes the image into a 
two-dimensional latent feature map $\mathbf{Q} \in \mathbb{R}^{h \times w \times n_z}$. The codebook is utilized to quantize the latent feature map by replacing each pixel embedding with its closest latent variables within the codebook element-wisely as follows:
\begin{equation}
    \mathbf{\hat{Q}}_{ij} = \arg\min_{\mathbf{z}_k}\parallel \mathbf{Q}_{ij} - \mathbf{z}_{k} \parallel^{2}.
\end{equation}
For reconstruction, the decoder $G$ takes the quantized latent feature map $\mathbf{\hat{Q}}$ as input and returns an generated image $\mathbf{\hat{X}}$ close to the original image, 
i.e., $\mathbf{\hat{X}} \approx \mathbf{X}$.

For image generation, the quantized feature map $\mathbf{\hat{Q}}$ can be modeled as a sequence of discrete tokens,
denoted as a sequence of discrete token indices $\mathbf{I} \in \{0, \dots, K-1\}^{h \times w}$.
Each token roughly corresponds to an image patch of the size $\frac{H}{h} \times \frac{W}{w}$. 
Thus, the prediction of a token sequence is equivalent to synthesizing an image. 
In practice, we refer to uni-directional Transformer\cite{vaswani2017attention} to predict the image sequence autoregressively as follows:
\begin{equation}
    P(\mathbf{I}_{\leq i} | \mathbf{T}) = \prod_{j}^{i}P(\mathbf{I}_j | \mathbf{I}_{<j}, \mathbf{T}),
\end{equation}
where $\mathbf{T}$ is the sequence of text tokens of the caption paired with image $\mathbf{X}$. 

To introduce positional information of the two modalities in Transformer, we learn two sets of positional embeddings. One is axial positional embeddings \cite{ho2019axial} for the visual sequence from a spatial grid. The other is sequence embeddings as BERT \cite{devlin2018bert} for text sequence. 

The autoregressive task minimizes cross entropy losses applied to the reconstruction of text tokens and image tokens, respectively~\cite{dalle}, 
% following DALL$\cdot$E\cite{dalle}:
\begin{equation}\label{l_txt}
    \mathcal{L}_{txt} = - \mathbb{E}_{\mathbf{T_i}}\log P(\mathbf{T}_i|\mathbf{T}_{<i}),
\end{equation}
\vspace{-0.08in}
\begin{equation}
    \mathcal{L}_{img} = - \mathbb{E}_{\mathbf{I_i}}\log P(\mathbf{I}_i|\mathbf{I}_{<i}, \mathbf{T}).
\end{equation}

\begin{remark}
One consequent training idea is the masked sequence modeling by optimizing the loss for the paired text and image tokens. However, unlike most existing vision-and-language models \cite{tan2019lxmert,ni2021m3p,zhou2021uc2} taking detected regions as an image sequence, our model accepts patch sequence, which will be an inexact alignment with text. Moreover, fine-grained correspondences of image patches and attribute tokens are difficult to be aligned. For example, aligning ``a red crown'' and ``a red belly'' within the detected bird needs to precisely recognize not only the color but also the body parts. To avoid noisy training signals, we do not adopt masked sequence modeling for training.
\end{remark}

\subsection{Training with Language and Vision Guidance}

\noindent\textbf{Language Guidance.}
The transformer model determines the basic image tokens at the top level, and the autoencoder model holds the convolutional decoder complementing the texture in detail at the bottom level. Training these two models separately implies splitting the generation stream stiffly. To this end, in our Mani component, we propose a semantic loss for the token prediction not only considering the downstream decoding but also improving the ability to capture the relation between text and image.

The CLIP \cite{patashnik2021styleclip} is a vision-and-language representation learning model, trained with 400 million image-text pairs and has shown excellent visual-semantic alignment capability by achieving superb performance on the task of zero-shot image classification. It is optimized by a symmetric cross entropy loss over the cosine similarities of a batch of image and text embeddings. In this work, we leverage the strong model CLIP to guide our token prediction, through
\begin{equation}
    \mathcal{L}_{semantic} = 1 - D(G(\mathbf{\hat{I}}), \mathbf{T}),
\end{equation}
where $D$ is the cosine similarity between the CLIP embeddings of its two arguments, as shown in Fig.~ \ref{fig:framework}. Note that the gradient back-propagation is implemented by straight-through estimator \cite{bengio2013estimating}.

\noindent\textbf{Vision Guidance.}
With the text descriptions, our model can replace an entity with other specific entities. For only editing the appearance of an entity, we need to provide the model with the prior information of the original entities' shape. Specifically, we convert the image to grayscale and append the quantized grayscale image tokens $\mathbf{V} \in \{0, \dots, K-1\}^{h \times w}$ to the text sequence as another condition for the tokens to be manipulated. The grayscale image token sequence $\mathbf{V}$ shares the positional embeddings with the image token sequence $\mathbf{I}$, for the same modality and spatial positions. For the identities of vision guidance and input text token sequence, we append two special separation tokens [BOV] and [BOT] to the beginning of them respectively. We apply the cross entropy loss on the vision guidance tokens as well,
\begin{equation}
    \mathcal{L}_{gray} = - \mathbb{E}_{\mathbf{V_i}}\log P(\mathbf{V}_i|\mathbf{V}_{<i}).
\end{equation}
We randomly select 50\% samples to train with vision guidance. The total loss to train the transformer model is a combination of the four losses, which can be divided into two parts,
\begin{equation}
    \mathcal{L}_{ar} = \lambda_{1}\mathcal{L}_{img} + \lambda_{2}\mathcal{L}_{gray} + \lambda_{3}\mathcal{L}_{txt},
\end{equation}
\vspace{-0.08in}
\begin{equation}
    \mathcal{L}_{total} = \mathcal{L}_{ar} + \lambda_{4}\mathcal{L}_{semantic},
\end{equation}
where $\lambda_1, \lambda_2, \lambda_3$ and $\lambda_4$ are the balancing coefficients.

\subsection{Inference with Entity Guidance}
\label{sec:semantic_alignment_module}

We design a semantic alignment module to locate the image patches to be manipulated by input text automatically in the inference phase. 
The semantic alignment module is a two-step module, (1) to find the tokens of every entity and (2) to select the text-related entities to be manipulated, where each step bases on a strong existing model.

In the first step, we refer to entity segmentation \cite{entity_segmentation} to recognize each entity on the original image $\mathbf{X}$, as Fig.~\ref{fig:framework} shows. The segmentation is implemented on the original image size, and we use the bilinear interpolation to resize the binary mask map of each entity to the same size of latent feature map $\mathbf{Q}$. The pixels whose values are larger than 0 represent that the tokens at the same position belong to the entity. In our preliminary experiments, we compare the bilinear interpolation with max-pooling for finding the entity tokens. The max-pooling dilates the tokens for the bilinear interpolation, however, due to the stack of convolutions in the first stage, the receptive field of the tokens by max-pooling is beyond the entity area and overlaps with other entities'. Thus, we use the bilinear interpolation to map the segment mask and token mask for a more precise alignment.

In the second step, we set a text prompt word to select the relevant entities. We leverage the FILIP \cite{yao2021filip}, a CLIP-style model optimized by token level similarity, to calculate the similarities between image token and text token. For example, as Fig.~\ref{fig:framework} shows, we set ``bird'' as a prompt word to search the bird entities in the image, and then we average the similarities between tokens of each entity and the prompt word ``bird''. The entities whose similarities are higher than $\theta$ are the text-related entities. 
\begin{remark}
There is another prevalent way, word-patch alignment, to align a pair of text and image tokens, especially in many multi-modality transformer methods \cite{radford2021clip, yao2021filip}.  The word-patch alignment begins from the word tokens to sort the patch tokens, which takes the image patches separately and neglects the information of the entity tokens as a whole during the alignment. Thus the selected image tokens may well scatter within or around an entity area. Manipulating these scattered tokens gets a messy image, where the foreground stays while the background changes. A comparison between word-patch alignment and our semantic alignment method is included in Section \ref{semantic_alignment}.
\end{remark}

\section{Experiments}

We compare ManiTrans against  ManiGAN~\cite{2019ManiGAN} and Lightweight-GAN~\cite{2020Lightweight}.
Results of competitors are reproduced using the code/model released by the authors.

\noindent \textbf{Datasets}.
Following common practice, we benchmark ManiTrans on three public datasets, including  CUB \cite{cub200}, the Oxford \cite{oxford102} and the more complicated COCO \cite{mscoco} datasets. The statistics of these datasets are in the Appendix.
We preprocess these datasets as in~\cite{zhang2017stackgan,xu2018attngan}.

\begin{figure*}
\centering
\includegraphics[width=\textwidth]{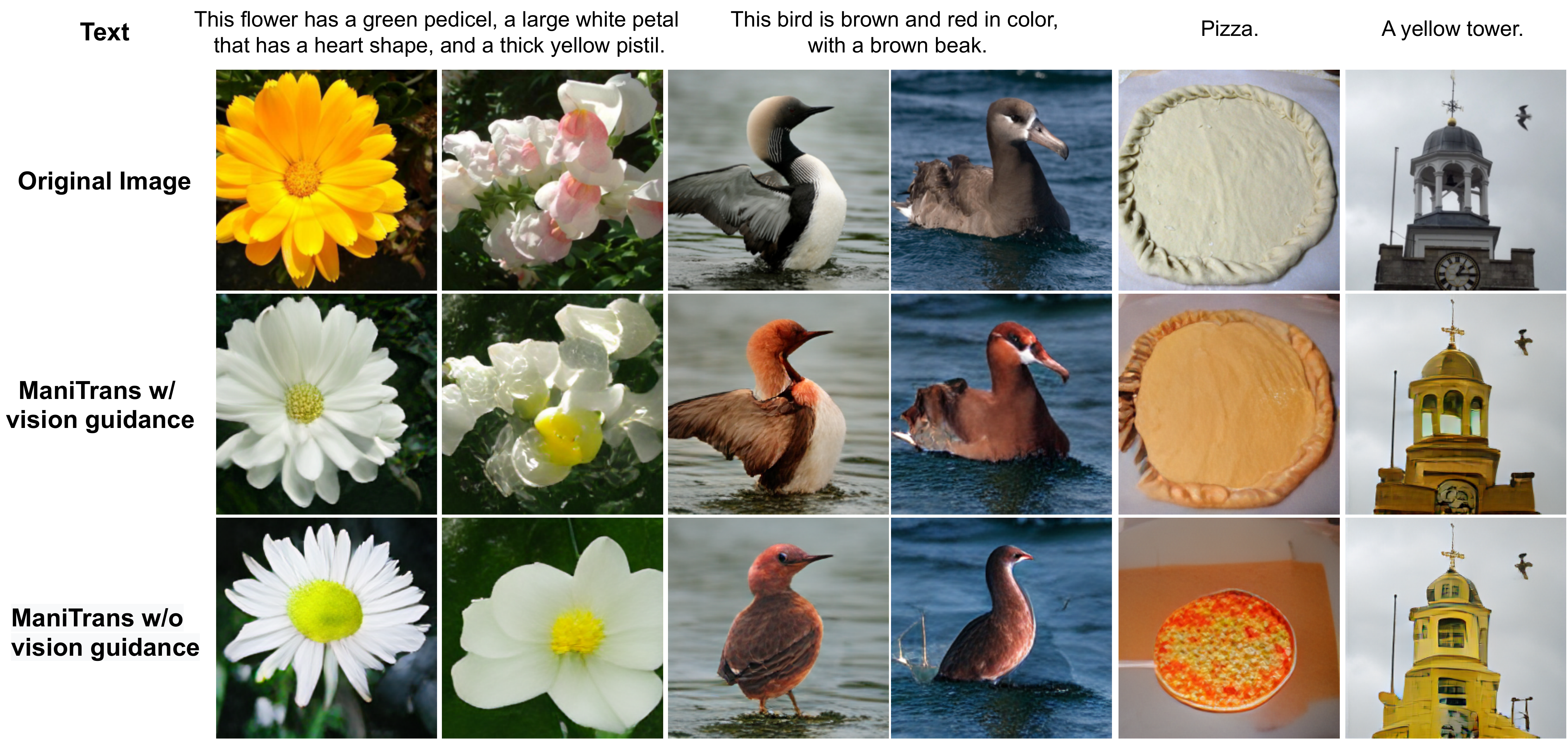}
% \vspace{-0.15in}
\caption{Our manipulation results w/ and w/o vision guidance on the Oxford, CUB and COCO datasets. With vision guidance, ManiTrans can well keep the structure of an object while without vision guidance, ManiTrans can flexibly generate an entity with a different structure according to the text. 
The prompt word for the text ``Pizza.'' is ``dough''.
}
\label{fig:main_results}
% \vspace{-0.15in}
\end{figure*}

\subsection{Quantitative Metrics}
% \footnote{https://github.com/toshas/torch-fidelity}
To evaluate the quality of manipulated images, we use the Inception Score (IS) \cite{salimans2016IS} as the quantitative evaluation metric. To evaluate the visual-semantic alignment between the text descriptions and manipulated images, we calculate the cosine similarity between their embeddings extracted with CLIP text/image encoders, called CLIP-score. 
Besides, we conduct an image-to-text retrieval experiment and report Recall@N for quantitative comparison. In the image-to-text retrieval, for each manipulated image, the text candidates consist of the input text, which serves as the positive sample, and 99 randomly sampled descriptions as negative samples. Such 100 text candidates are sorted in descending order according to their cosine similarity with the manipulated image. Recall@N calculates the percentage of images, whose positive sample occurs within the top-N candidates.
As we use the ViT-B/32 CLIP model during training, for a fair comparison, we refer to the ResNet50 CLIP model to compute the CLIP-score. 
Additionally, following \cite{tagan}, to compare the quality of the content preservation, we compute the L2 reconstruction error by forwarding images with positive texts. 

The higher the IS, the higher quality of the manipulated images. Higher CLIP score and R@N indicate better visual-semantic alignment between the input texts and the manipulated images. The lower the L2 error, the higher content preservation quality.

\begin{figure*}
\centering
\includegraphics[width=\textwidth]{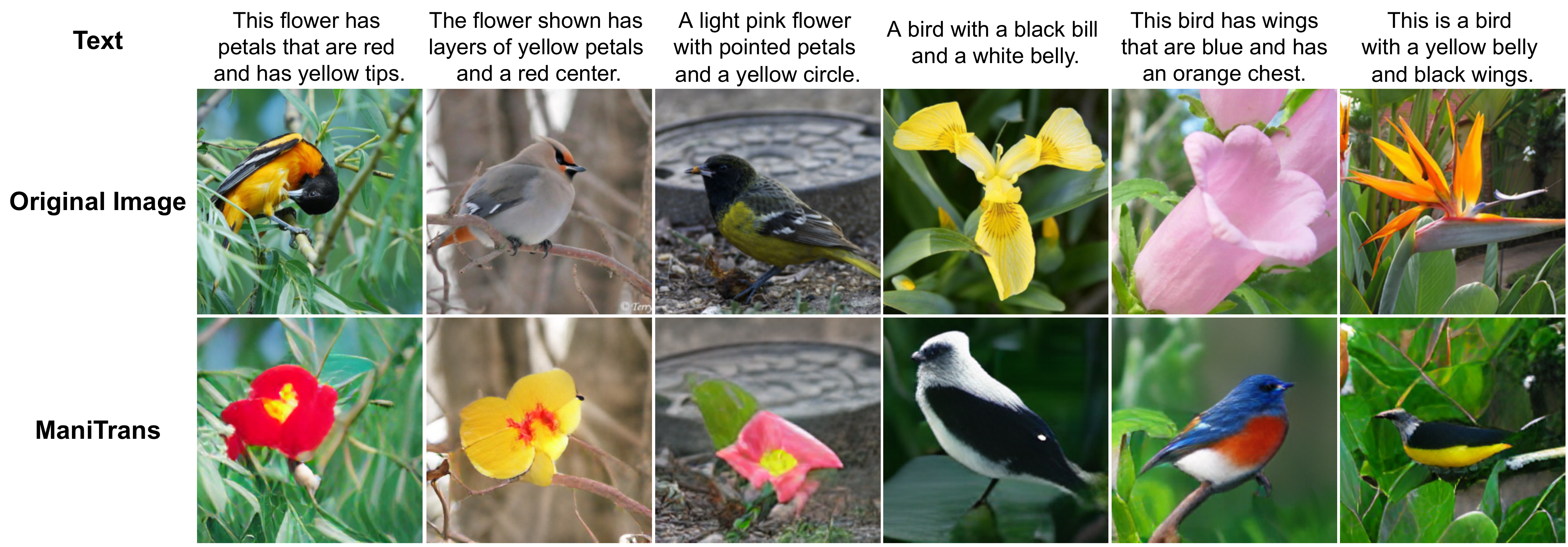}
% \vspace{-0.15in}
\caption{Manipulation results from bird to flower and flower to bird with our proposed ManiTrans.\label{fig:birdflowerexchange}}

\end{figure*}

\begin{table*}
\centering
\resizebox{\textwidth}{!}{
\begin{tabular}{lcccccccccccc}
\hline
\multirow{2}{*}{Model} & \multicolumn{4}{c}{CUB}                    & \multicolumn{4}{c}{Oxford}                 & \multicolumn{4}{c}{COCO}                    \\  \cmidrule(r){2-5}\cmidrule(r){6-9}\cmidrule(r){10-13}
                       & IS        & CLIP-score & R@10   & L2-error & IS        & CLIP-score & R@10   & L2-error & IS         & CLIP-score & R@10   & L2-error \\ \midrule
ManiGAN                & 4.19 \small{$\pm$}  0.04 & 21.30     & 10.49 & 0.05    & 4.37 \small{$\pm$}  0.11 & 21.59     & 14.21 & \textbf{0.02}    & 22.6 \small{$\pm$}  0.40 & 11.91     & 14.50 & 0.03    \\\midrule
Lightweight-GAN        & 4.66 \small{$\pm$}  0.06 & 18.88     & 10.00 & 0.13    &    4.35 \small{$\pm$} 0.09       & 17.55    & 11.58 & 0.12    & \textbf{24.80 \small{$\pm$}  0.94} & \textbf{13.65}     & 14.49 & 0.03    \\\midrule
Ours                   & \textbf{5.02 \small{$\pm$}  0.11} & \textbf{23.56}     & \textbf{34.82} & \textbf{0.01}    & \textbf{4.50 \small{$\pm$}  0.06} & \textbf{23.34}   & \textbf{36.49} & 0.03   & 21.45 \small{$\pm$}  0.41 & 13.10    & \textbf{21.32} & \textbf{0.02}   \\ \hline
\end{tabular}
}
% \vspace{-0.1in}
\caption{Quantitative comparison between ManiGAN \cite{2019ManiGAN}, Lightweight-GAN \cite{2020Lightweight} and our ManiTrans. IS - Inception Score, CLIP-score - averaged cosine similarity with CLIP embeddings, R@10 - recall within top 10 candidates, L2-error - L2 reconstruction error. IS, CLIP-score and R@10 are the higher the better, L2-error is the lower the better.\label{table:main-results}}

% \vspace{-0.15in}
\end{table*}

\begin{figure*}
\centering
\includegraphics[width=0.96\textwidth]{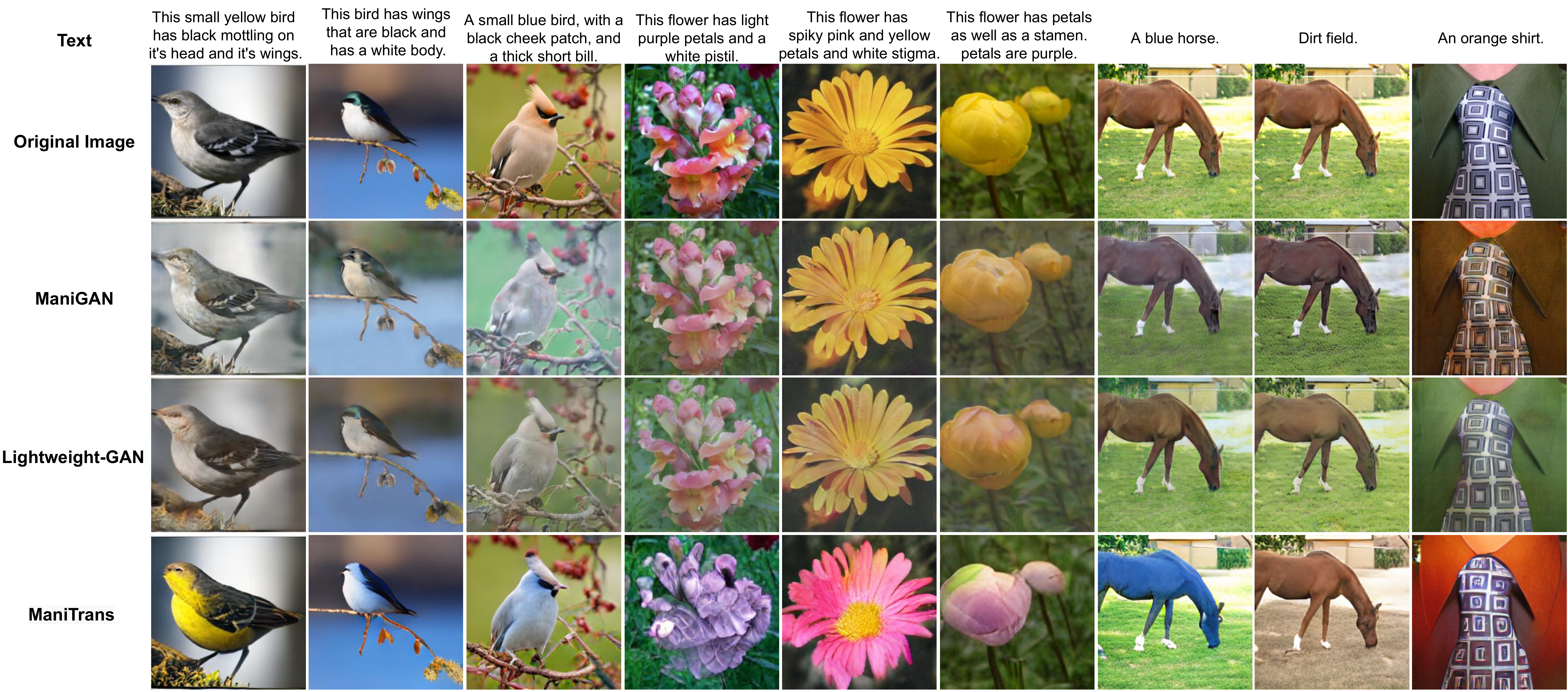}
% \vspace{-0.15in}
\caption{Qualitative comparison of different methods on the CUB, Oxford and COCO datasets. For a fair comparison, our ManiTrans uses the vision guidance to manipulate the images.}
\label{fig:comparewithgan}
% \vspace{-0.15in}
\end{figure*}

\subsection{Experimental Setup}

The model at the first stage inherits from the VQGAN \cite{vqgan} pretrained on ImageNet, where the codebook size is 1024, the image size is $256 \times 256$, and the latent feature map size is $16 \times 16$. At the second stage, our transformer has 24 layers, 8 heads with 64 dimensionalities for each head. We replace the traditional Feed-Forward Network (FFN) with a GEGLU \cite{shazeer2020glu} variant, which adds a Gated Linear Units (GLU) \cite{dauphin2017language} with GELU \cite{hendrycks2016gaussian} activation to the first hidden layer of FFN. 
We use Byte-Pair Encoding \cite{sennrich2015neural} to tokenize the text, with vocabulary size 49408. We limit the text length to 128 and learn a padding token for each position as DALL$\cdot$E.
Our transformer has 152M parameters, a little larger than BERT-Base 110M. The hyper-parameters of autoregressive loss $\lambda_1, \lambda_2, \lambda_3$ are set to $7/9, 1/9, 1/9$ and $\lambda_4$ of language guidance loss is $5$ for all the datasets. 
The CLIP model for the semantic loss is ViT-B/32
\footnote{https://github.com/openai/CLIP}
. 
For the semantic alignment module, we use the entity segmentation model based on Swin-L-W7 
\footnote{https://github.com/dvlab-research/Entity} 
and the FILIP-large\cite{yao2021filip} model for similarity computation.  The similarity threshold $\theta$ is 0.163.

For a good initialization of the transformer, we pretrain our transformer on CC12M \cite{changpinyo2021cc12m} without language and vision guidance. We use AdamW optimizer with $\beta_1=0.9, \beta_2=0.96$ to train 12 epochs with batch size 112. The learning rate linearly ramps up to $6 \times 10^{-4}$ for the first 5k iterations and is halved whenever training loss does not decrease for 50000 iterations. With the same optimizer, we fine-tune our model on the three datasets with the same two steps. The first step fine-tunes the model without vision guidance. The second step adds the vision guidance into training with 50\% samples. Each step lasts 500 epochs with batch size 96 and the learning rate linearly ramps up to $5 \times 10^{-4}$ for the first 1k iterations and is halved when training loss does not improve for 10 epochs.
We implemented the proposed ManiTrans with Huawei MindSpore\cite{mindspore} and Pytorch; both implementations show comparable performance and efficiency.

% The codes and models will be released. 
As discussed in Section~\ref{sec:semantic_alignment_module}, we set a prompt for the entity to be manipulated in the inference phase. Particularly, CUB and Oxford have specific category images, where we set ``bird" and ``flower" as the prompt word respectively. COCO contains various category entities, and we set a prompt word for each text guide of each image. Almost all the prompt words of COCO are the nouns of their text in the following experiments and we will state the prompt words for special examples.

\subsection{Main Results}

In this section, we first qualitatively verify the manipulation ability of our model to edit or change the entity on the Oxford, CUB and COCO datasets. As Fig.~\ref{fig:main_results} shows, our model can manipulate the images with the same object structure providing the vision guidance, i.e. the grayscale image, as prior shape information. Without the constraint of the vision guidance, our model generates an apparently different entity corresponding to the text description in the place of the original entity. Our framework merges the generation ability to the manipulation without any user manual mask but only the guidance of input text, where most existing models fail. 

We also conduct an experiment trained on the mixture of datasets, CUB and Oxford, to verify a wider manipulation than on the same category. 
As shown in Fig.~\ref{fig:birdflowerexchange}, ManiTrans generates reasonable manipulated entities which are corresponding to the text and fit to the background, in both bird-to-flower and flower-to-bird settings. For example, 
in the third column from the left, ManiTrans not only generates a flower consistent with the description but also complements the upper left corner of the manipulated flower with a leaf, 
which shows that ManiTrans also learns a combination of the object information and the background. 
Please refer to the supplementary material for more results.

\subsection{Comparison with the State of the Art}
Table \ref{table:main-results} shows the quantitative comparison of our method against previous methods, including 
ManiGAN \cite{2019ManiGAN} and Lightweight-GAN \cite{2020Lightweight}. 
On CUB and Oxford datasets, our ManiTrans achieves better results than other models on almost all metrics, except for the L2-error on Oxford datset, where ManiTrans is competitive with ManiGAN.
It demonstrates that our method can generate high-quality manipulated images (IS), which are consistent with the text descriptions (CLIP-score and R@10) 
, and preserve the content of original images (L2-error).

For the more complicated dataset, COCO, ManiTrans outperforms the ManiGAN and Lightweight-GAN on the R@10 and L2-error and achieves competitive CLIP-score. The IS of our method are competitive with ManiGAN and Lightweight-GAN.  
However, as Fig.~\ref{fig:comparewithgan} shows, 
within many text-guided manipulation cases, ManiGAN and Lightweight-GAN both change the images slightly, more like applying a filter, while ManiTrans conducts manipulation according to the text. Typically, the former one is easier to generate high quality images than the latter
and this is why their IS are a bit higher than our method. 

As Fig.~\ref{fig:comparewithgan} shows, 
compared with the original images, ManiGAN directs the images toward the semantic of text closer than Lightweight-GAN but changes the background style further from the original as well. Lightweight-GAN preserves the irrelevant contents better than ManiGAN while failing in transforming the text-relevant regions according to the descriptions. Our method outperforms them both on background preservation and foreground manipulation. As the 
% last 
second and third columns from the right of the Fig.~\ref{fig:comparewithgan} show, our ManiTrans can manipulate the horse and the field respectively on one image, while the baseline methods only change the whole image style with the text.

\begin{figure}
\centering
\includegraphics[width=\linewidth]{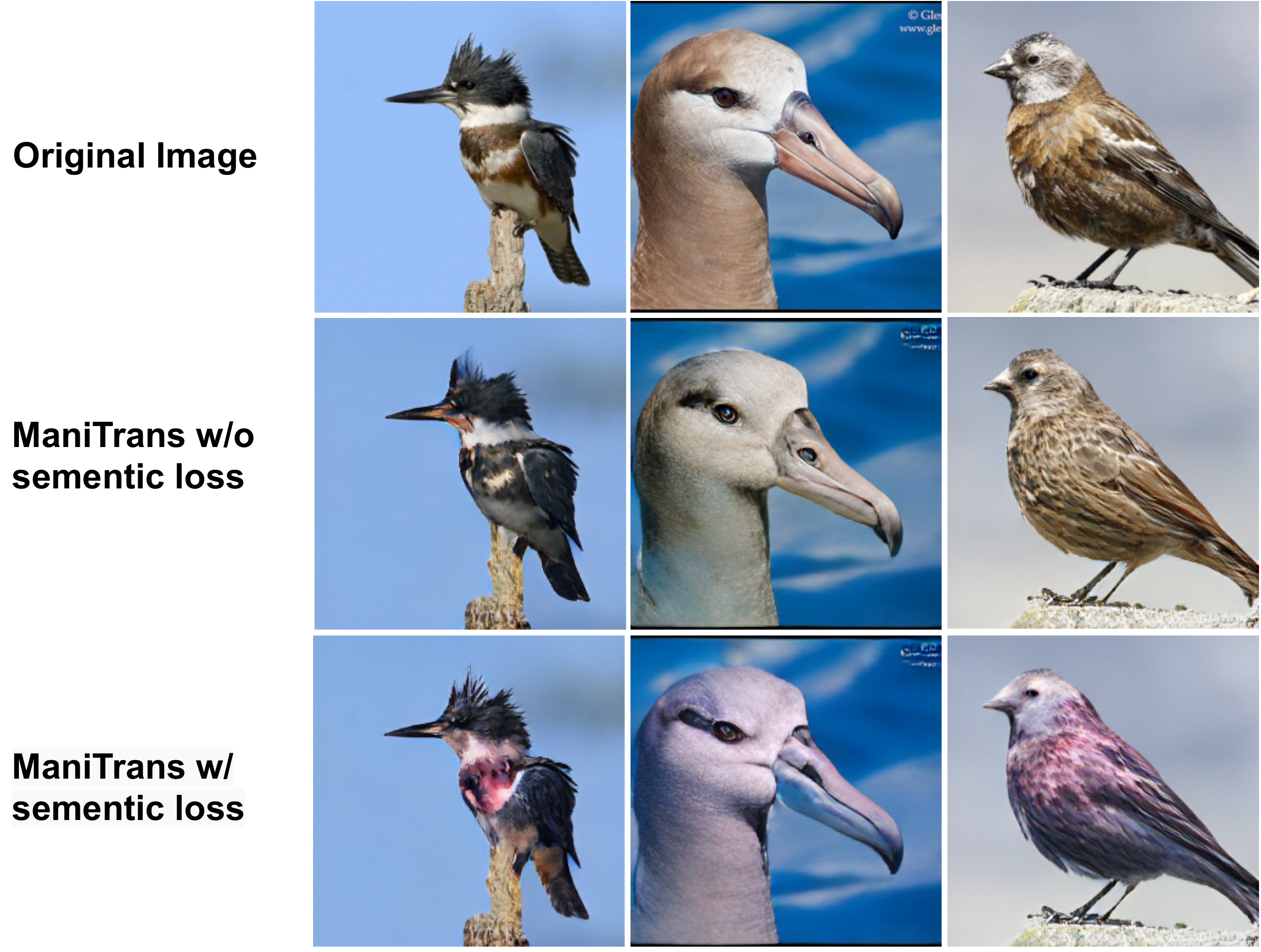}
% \includegraphics[scale=0.3]{figures/CLIP_loss.pdf}
% \vspace{-0.25in}
\caption{Qualitative comparison of our methods w/ and w/o semantic loss on CUB dataset. The input text is ``This particular bird has a belly that is purple and gray.''.}
\label{fig:semantic}
% \vspace{-0.15in}
\end{figure}

\noindent \textbf{Effects of Semantic Loss}.\label{cliploss}
Fig.~\ref{fig:semantic} compares the qualitative results of our model with the semantic loss or not on CUB dataset. The model trained with the semantic loss manipulates the bird as gray and purple, while the model trained without the semantic loss neglects the purple. It implies the semantic loss helps the model capture the relation between image and text. 

\begin{figure}
\centering
\includegraphics[width=\linewidth]{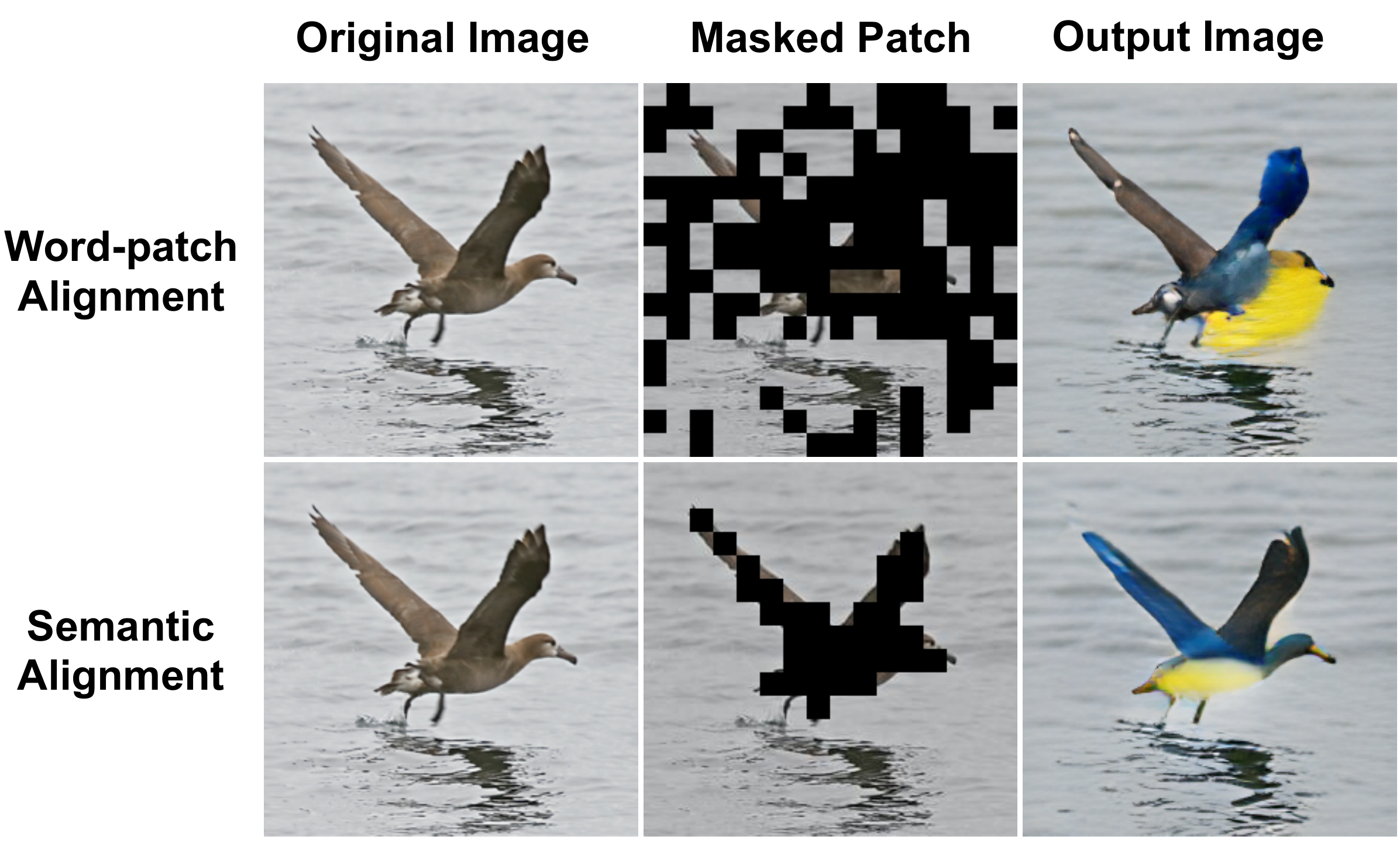}
% \vspace{-0.25in}
\caption{Qualitative comparison of our methods with our semantic alignment mechanism and word-to-patch alignment on CUB dataset. The input text is ``This bird has wings that are blue and has a yellow belly.''.}
\label{fig:align}
% \vspace{-0.10in}
\end{figure}

\noindent \textbf{Effects of Semantic Alignment Module}.
\label{semantic_alignment}
We compare our semantic alignment against the word-patch alignment qualitatively in Fig.~\ref{fig:align}. The two methods share the same similarity threshold $\theta$ for sorting the image tokens.
% to be manipulated. 
As Fig.~\ref{fig:align} shows, our semantic alignment selects the image patches corresponding to the bird precisely, while the word-patch alignment misses some patches corresponding to the bird and selects a few patches which belong to the background. With the inaccurate patches selected by word-patch alignment, only the right-wing turns to blue and the yellow leaks out. Although the color of two manipulated images both match the description, the qualitative result by the semantic alignment is better, resulting from more precise edited locations.

\begin{figure}
\centering
\includegraphics[width=\linewidth]{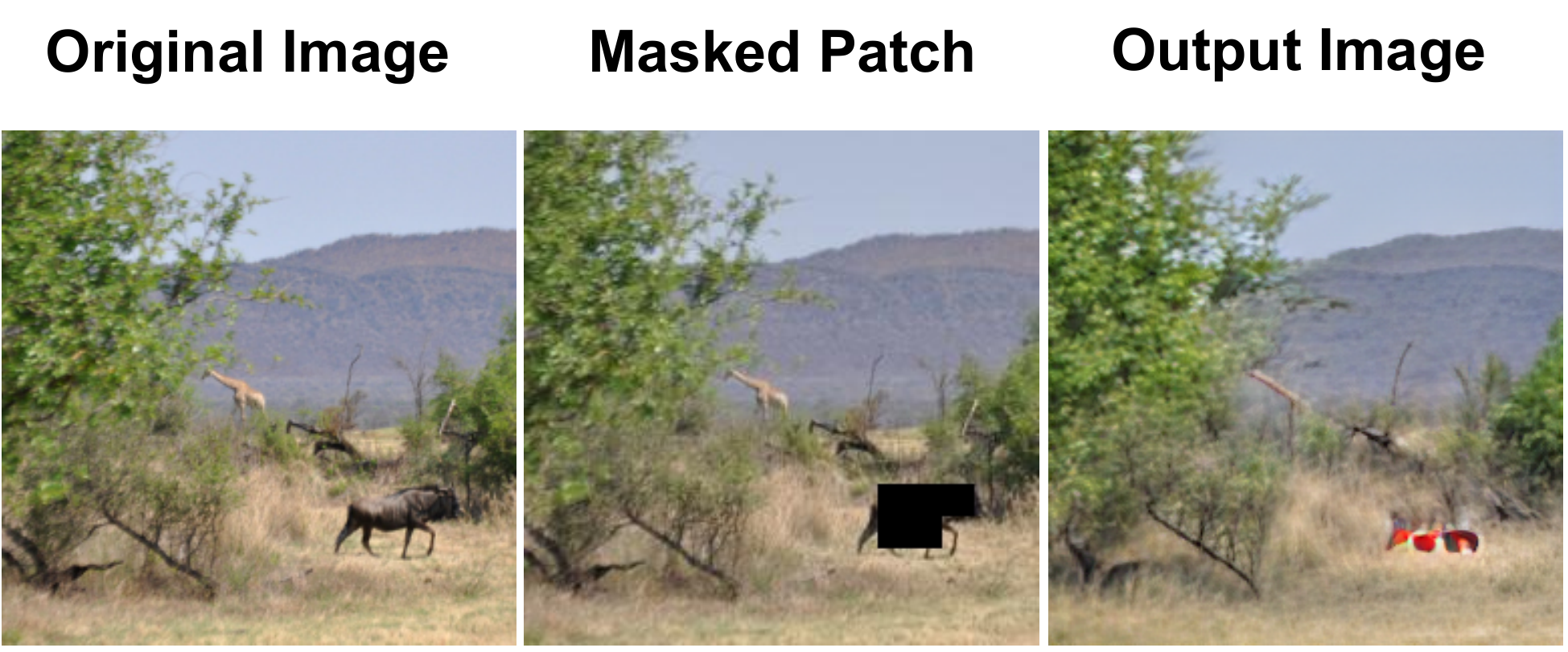}
% \vspace{-0.25in}
\caption{A failed manipulated example on COCO dataset. The input text is ``A red giraffe.''.}
\label{fig:failure}
% \vspace{-0.15in}
\end{figure}

\noindent \textbf{Failed Case}.
Our method relies on a pretrained entity segmentation and FILIP model to locate the entities to be manipulated. 
Though the text prompt facilitates the model to locate different entities on one image, entities with similar appearances may be unexpectedly confused by the semantic alignment model.
We give a failed example in  Fig.~\ref{fig:failure}, where the semantic alignment module takes the wildebeest as the relative entity to ``giraffe''. Consequently, we fail the manipulation with red patches in the wrong place. 

\section{Conclusion}

For the first time, this paper studies a new task -- entity-level text-guided image manipulation. To tackle this task, we propose  a novel framework -- ManiTrans of Manipulating
Transformers with the token-wise semantic alignment and generation. It is the two-stage framework with a semantic alignment module to manipulate the images on the entity level, which incorporates editing and generation. 

\noindent \textbf{Social Impact}. Our ManiTrans offers a new tool to modify the images. 
Our framework can be used for many industrial applications on  education, and potentially help the visually impaired people. We do not expect the negative social impact, as the synthesized images should be generated with the textual guidance.

\noindent \textbf{Acknowledgement} {\small This work was supported in part by NSFC Project
(62176061), and SMSTM Projects (2018SHZDZX01 and 2021SHZDZX0103). The corresponding authors are Yanwei Fu, and Hang Xu.}

\newpage
%%%%%%%%% REFERENCES
{\small
\bibliographystyle{ieee_fullname} 
\bibliography{egbib}
}

\newpage
% % \twocolumn
% \begin{mytitlepage}
% \title{\textbf{ManiTrans: Entity-Level Text-Guided Image Manipulation via Token-wise
% Semantic Alignment and Generation: Supplementary Material}}
% \maketitle
\clearpage

\appendix
\renewcommand{\appendixname}{Appendix~\Alph{section}}
  
In this supplementary material, we provide the statistics of the datasets, quantitative results of ablation study, and qualitative results to further demonstrate the ability of our ManiTrans, and to discuss failed cases.

\section{Experiment Datasets}

As we mention in the main paper, we benchmark ManiTrans on three public datasets, including the CUB\cite{cub200},
Oxford \cite{oxford102} 
and COCO\cite{mscoco} 
datasets.
The number of images and the number of captions per image of each dataset are listed in Table \ref{tab:statistic_datasets}. The CUB and Oxford are two datasets about birds and flowers respectively. While there are at least 80 categories of objects with different shape structures and appearances on COCO images.
Thus, COCO is a more complicated dataset than CUB and Oxford, not only in model understanding the correspondence between the image and text, but also in image manipulation on the entity level.

\begin{table}[h]
\centering
\begin{tabular}{ll|c|c}
\hline
\multicolumn{2}{c|}{Dataset}                             & \#images & \#captions/image    \\ \hline
\multicolumn{1}{l|}{\multirow{2}{*}{CUB\cite{cub200}}}    & train & 8855      & \multirow{2}{*}{10} \\ \cline{2-3}
\multicolumn{1}{l|}{}                            & test  & 2933      &                     \\ \hline
\multicolumn{1}{l|}{\multirow{2}{*}{Oxford\cite{oxford102}}} & train & 7034      & \multirow{2}{*}{10} \\ \cline{2-3}
\multicolumn{1}{l|}{}                            & test  & 1155      &                     \\ \hline
\multicolumn{1}{l|}{\multirow{2}{*}{COCO\cite{mscoco}}}     & train & 80k       & \multirow{2}{*}{5}  \\ \cline{2-3}
\multicolumn{1}{l|}{}                            & test  & 40k       &                     \\ \hline
\end{tabular}
\caption{Statistics of datasets.}
\label{tab:statistic_datasets}
\end{table}

\section{Quantitative Results of Ablation Study}

\noindent\textbf{Effects of Vision Guidance.} 
% Thanks. We clarify that, as discussed in L372-376, 
Vision guidance aims to provide prior structures of entities and to make our model better generate the appearance of entities.  
Without such prior from vision guidance, our model generates entirely different entities, such as shown in Fig.~\ref{fig:supp_cub} - Fig.~\ref{fig:supp_exchange}. 
In Table~\ref{tab:visionguidance}, we present the quantitative results of our ManiTrans with and without vision guidance on CUB. Two semantic metrics, CLIP-score and R@10, are higher without vision guidance. The other two image quality metrics, IS and L2-error, are still competitive without the prior information by vision guidance.
% We clarify that even w/o vision guidance, our results are still competitive quality and better semantic of images on CUB, as Tab.~\ref{visionguidance} shows.

\noindent\textbf{Effects of Semantic Loss.} Table~\ref{tab:ablation} shows the quantitative results of the ablation study on semantic loss (SL). 
With SL, our ManiTrans achieves a higher CLIP-score and R@10, which demonstrates that SL helps the model improve the relevance between the manipulated images and text guides. 

\noindent\textbf{Effects of Semantic Alignment Module.}
Table~\ref{tab:ablation} compares the quantitative metrics when we apply the semantic alignment module (SAM) in the inference phase or not (\emph{i.e.} word-path alignment). All the semantic metrics (CLIP-score, R@10) and image quality metrics (IS, L2-error) are better with SAM. This suggests the manipulated images by SAM are more realistic and more consistent with the text semantics.

\begin{table}[]
\centering
\resizebox{\linewidth}{!}{
\begin{tabular}{ccccc}
 \toprule
Vision Guidance & IS & CLIP-score & R@10  & L2-error \\ \midrule
\checkmark & \textbf{5.02 \small{$\pm$} 0.11} & 23.56 & 34.82 & \textbf{0.01}    \\\midrule
-- & 4.98 \small{$\pm$} 0.06 & \textbf{24.02} & \textbf{42.61} & 0.02    \\
 \bottomrule
\end{tabular}
}
\caption{Quantitative results on CUB w/ or w/o vision guidance.}
 \label{tab:visionguidance}
\end{table}

\begin{table}[]
\centering
\resizebox{\linewidth}{!}{
\begin{tabular}{lccccc}
 \toprule
SL & SAM & IS & CLIP-score & R@10  & L2-error \\ \midrule
\checkmark & \checkmark & \textbf{5.02 \small{$\pm$} 0.11} & \textbf{23.56} & \textbf{34.82} & \textbf{0.01}    \\\midrule
\checkmark & - & 4.93 \small{$\pm$} 0.08 & 23.53 & 32.89 &  0.01   \\\midrule
- & \checkmark & 5.01 \small{$\pm$} 0.09 & 22.19 & 16.06 & 0.01  \\
 \bottomrule
\end{tabular}
}
 \caption{Quantitative results  on CUB. ``SL'' and ``SAM'' are for semantic loss and  semantic alignment module, respectively.}
 \label{tab:ablation}
\end{table}

\section{Comparison with StyleCLIP}
\begin{figure*}
\centering
\includegraphics[width=\textwidth]{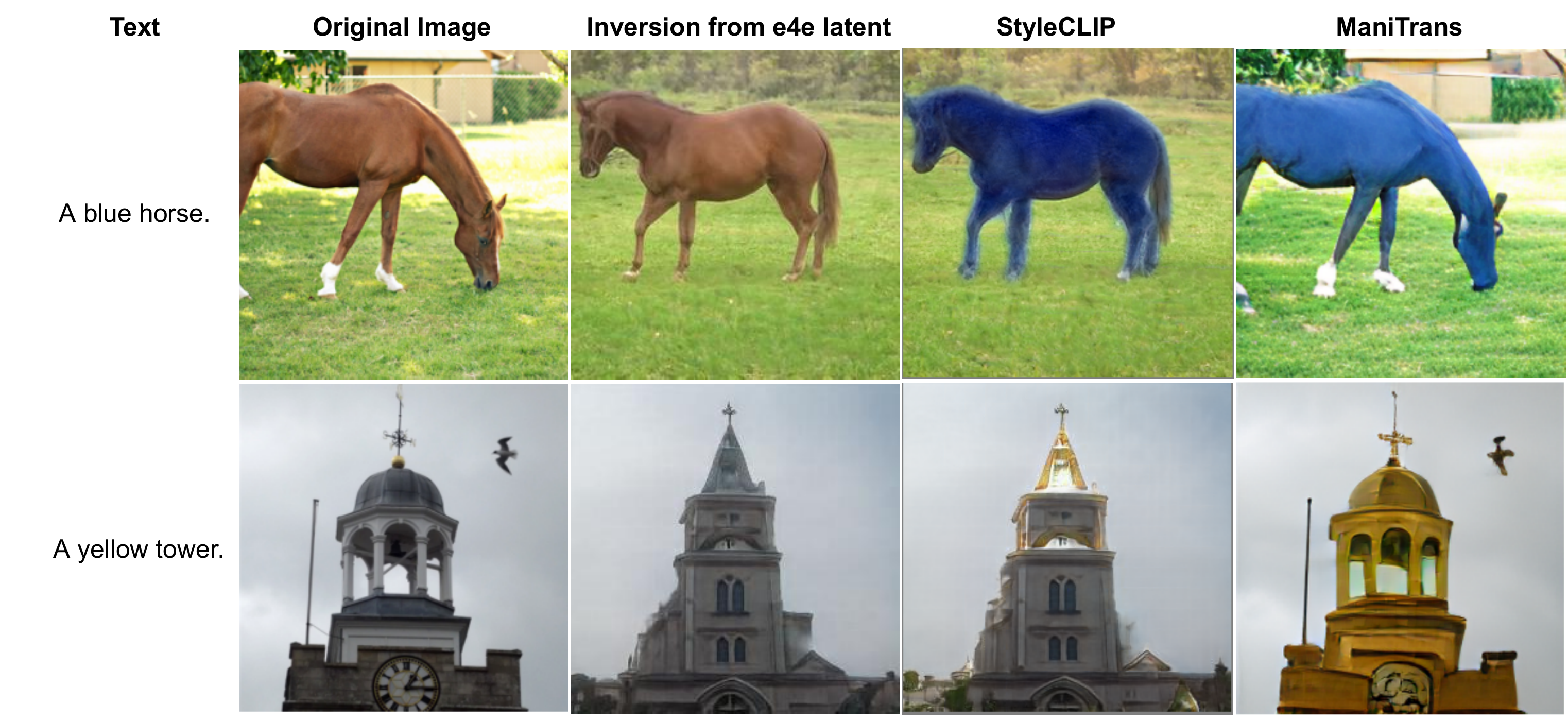}
\caption{Comparision with StyleCLIP.}
\label{fig:styleclip}
\end{figure*}

StyleCLIP \cite{patashnik2021styleclip} is one pioneer work on style transfer. However, StyleCLIP is more focused on faces, rather than nature images in our paper. 
In Fig.~\ref{fig:styleclip}, we provide a qualitative comparison with the global direction method of StyleCLIP, whose results are generated with official codes \& models from StyleCLIP and StyleGAN2.
StyleCLIP edits the latent of image inverted from e4e\cite{e4e}, whose content is different from the original image as shown in Fig.~\ref{fig:styleclip}. Furthermore, we achieve better manipulation on the tower.

% Previous works, such as StyleCLIP, are designed for manipulating the appearance of the one entity in an image. Our method aims to manipulate multiple entities for appearance and across different categories. It is not immediately obvious to us to conduct a fair comparison with StyleCLIP. Fig. {} shows ...

\section{Additional Qualitative Results}
\begin{figure*}
\centering
\includegraphics[width=\textwidth]{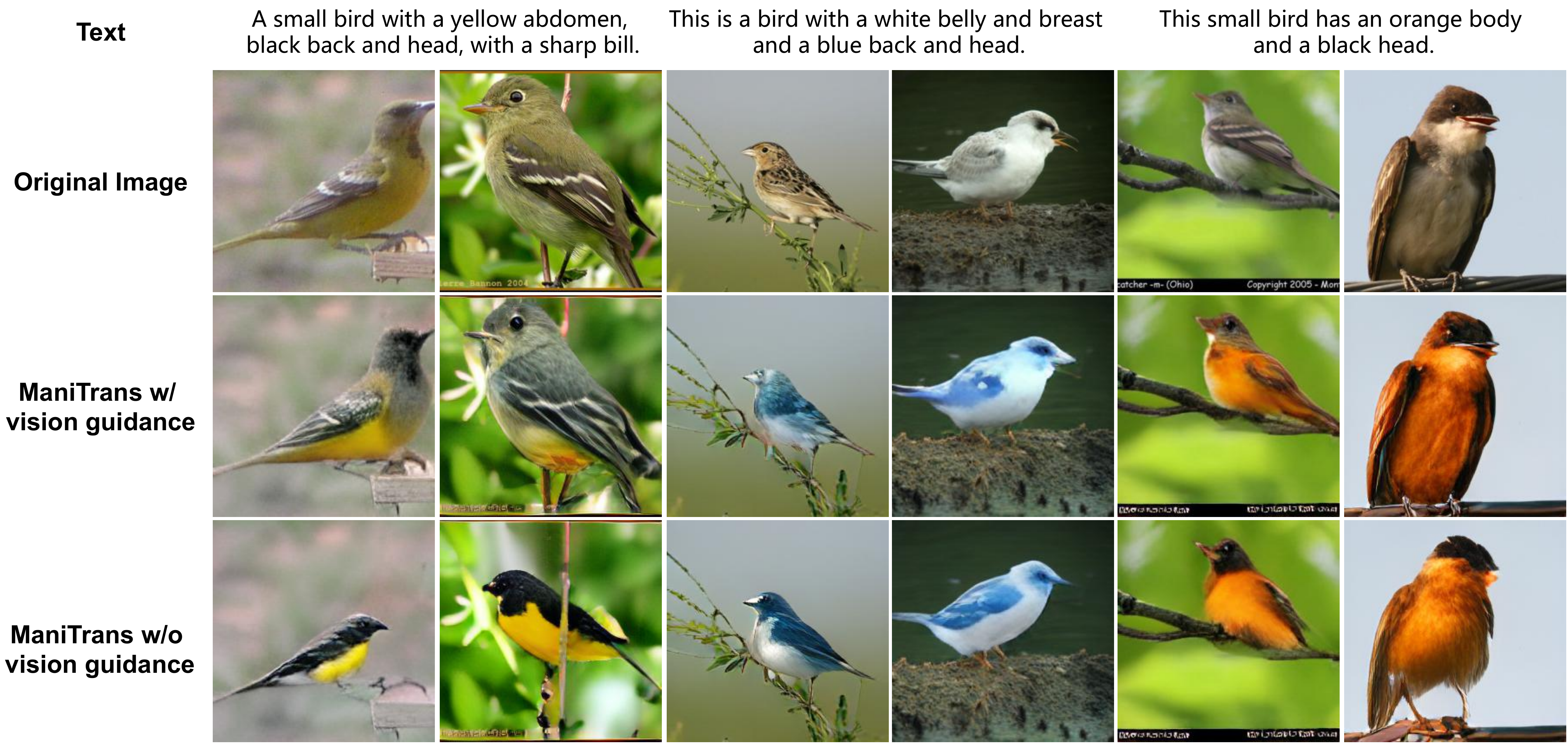}
\vspace{-0.15in}
\caption{Our manipulation results w/ and w/o vision guidance on CUB.}
\label{fig:supp_cub}
% \vspace{-0.15in}
\end{figure*}

\begin{figure*}
\centering
\includegraphics[width=\textwidth]{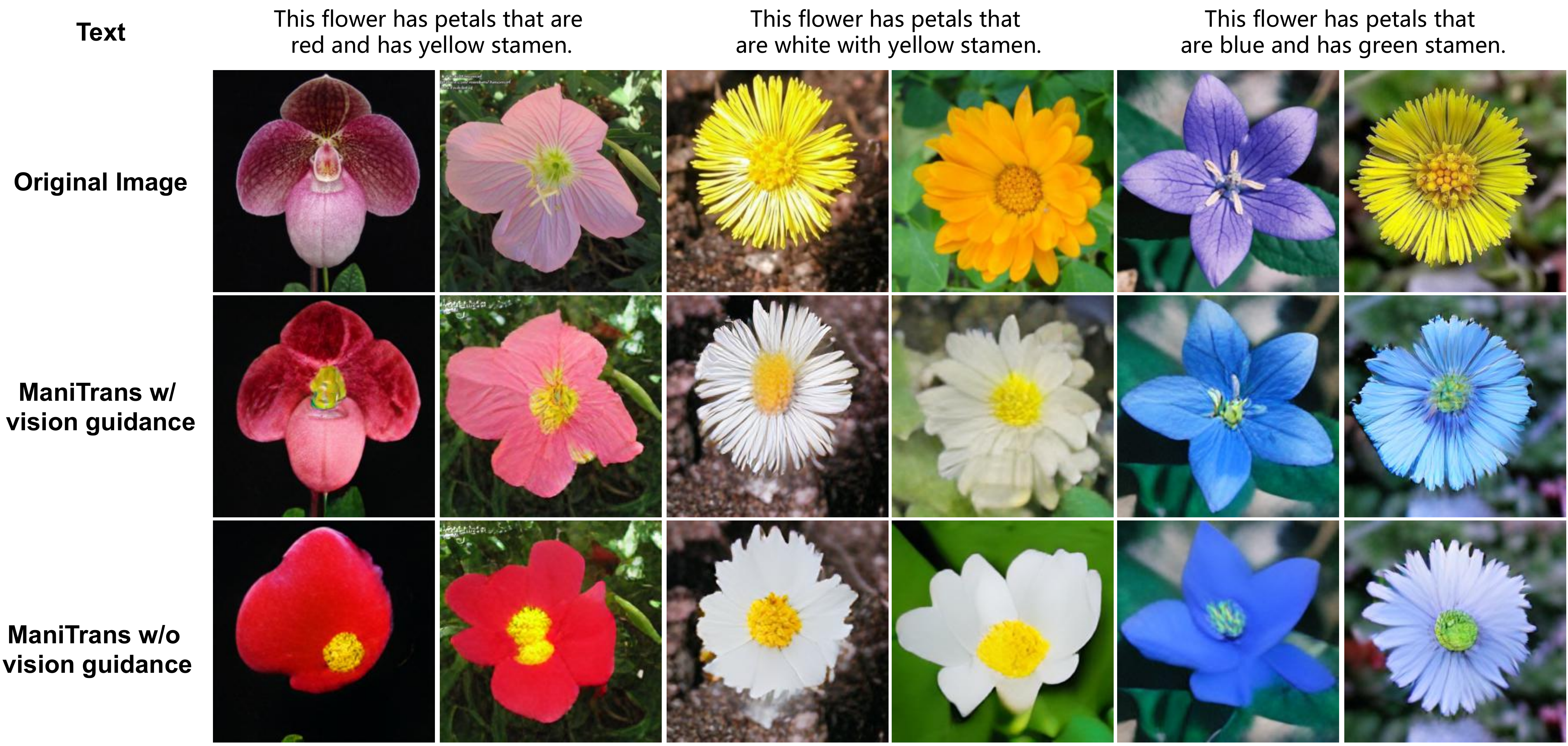}
\vspace{-0.15in}
\caption{Our manipulation results w/ and w/o vision guidance on Oxford.}
\label{fig:supp_oxford}
% \vspace{-0.15in}
\end{figure*}

\begin{figure*}
\centering
\includegraphics[width=\textwidth]{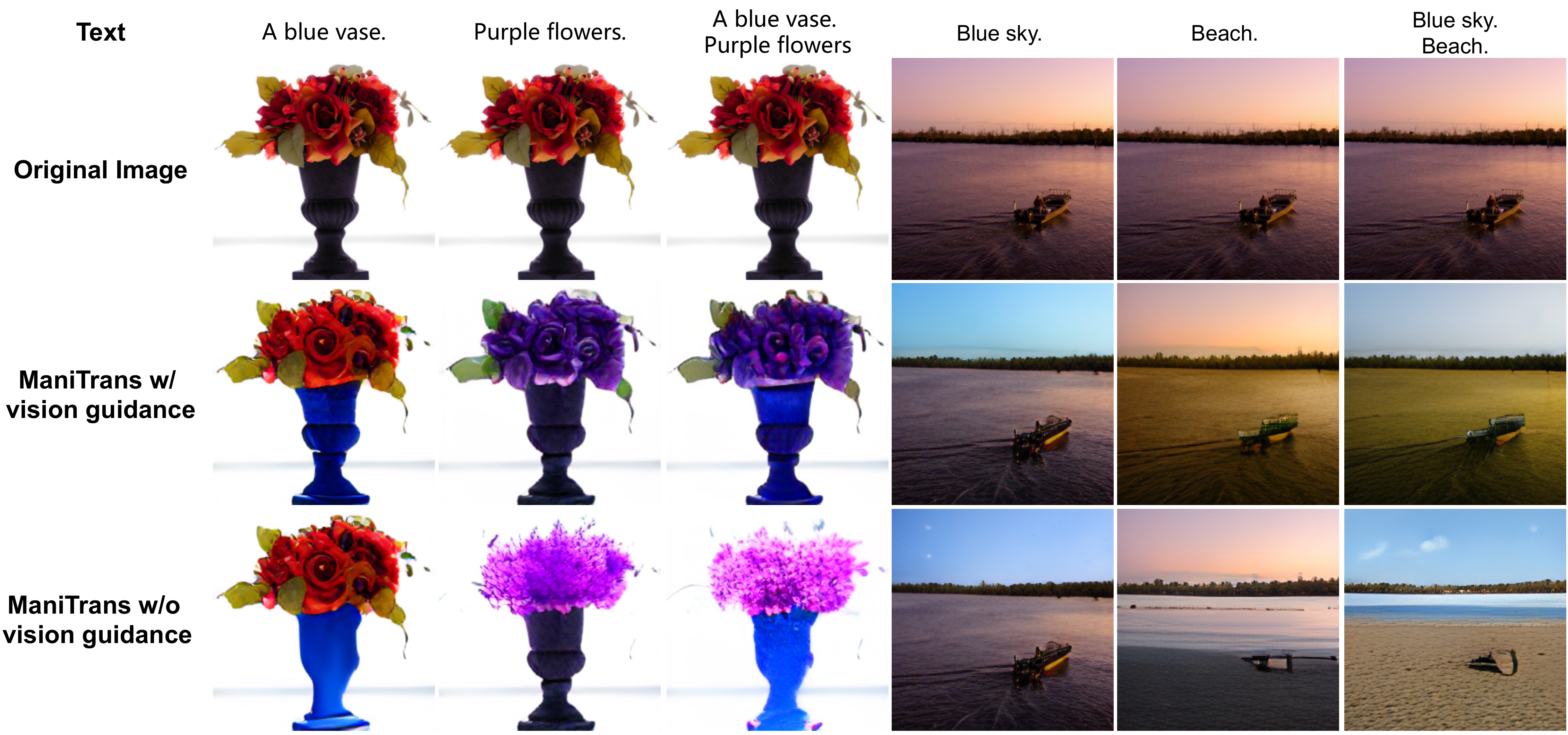}
% \vspace{-0.15in}
\caption{Our manipulation results w/ and w/o vision guidance on COCO. Our ManiTrans can manipulate different entities separately and together on one image. The prompt word for the text ``Beach.'' is ``river''.}
\label{fig:supp_coco}
% \vspace{-0.15in}
\end{figure*}

\begin{figure*}
\centering
\includegraphics[width=\textwidth]{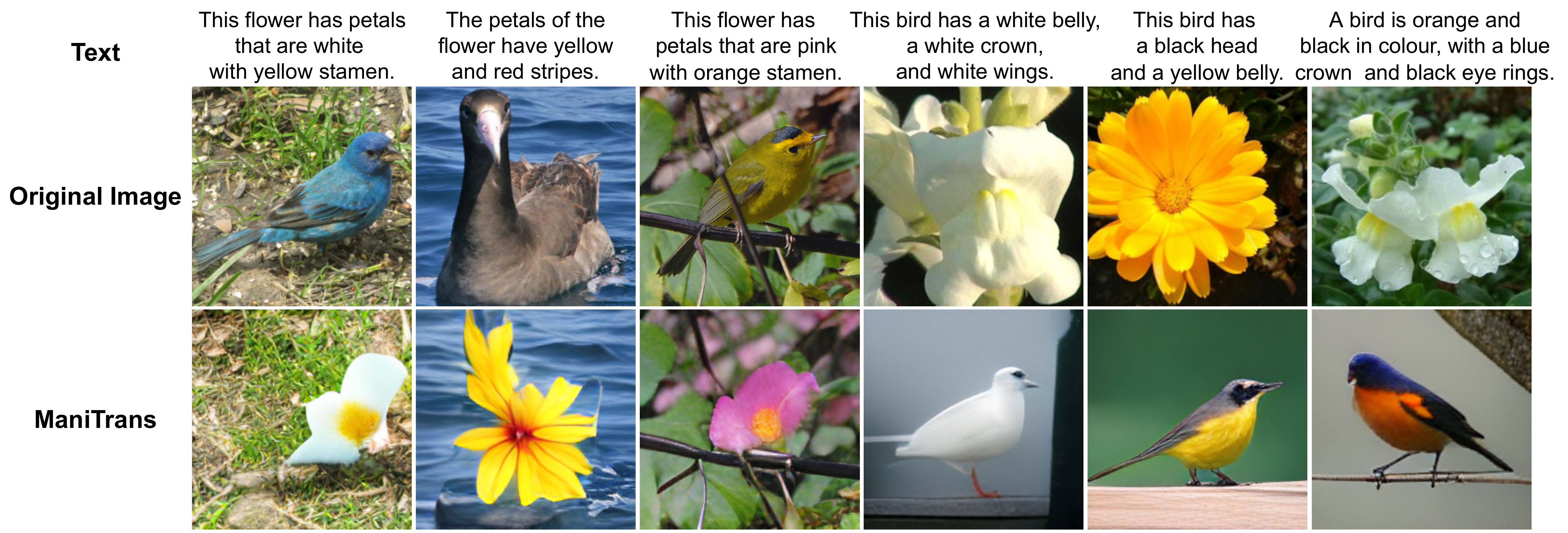}
\vspace{-0.15in}
\caption{Our manipulation results from bird to flower and flower to bird.}
\label{fig:supp_exchange}
% \vspace{-0.15in}
\end{figure*}

We provide more qualitative results on CUB (Fig.~\ref{fig:supp_cub}), Oxford (Fig.~\ref{fig:supp_oxford}), COCO (Fig.~\ref{fig:supp_coco})
, and a bi-directional entity transformation between bird and flower (Fig.~\ref{fig:supp_exchange}).
The prompt words of CUB and Oxford are ``bird" and ``flower". Almost all the prompt words of COCO are the nouns of their text and we will state the prompt words for special examples.

\section{Additional Failed Cases}

\begin{figure*}
\centering
\includegraphics[width=\textwidth]{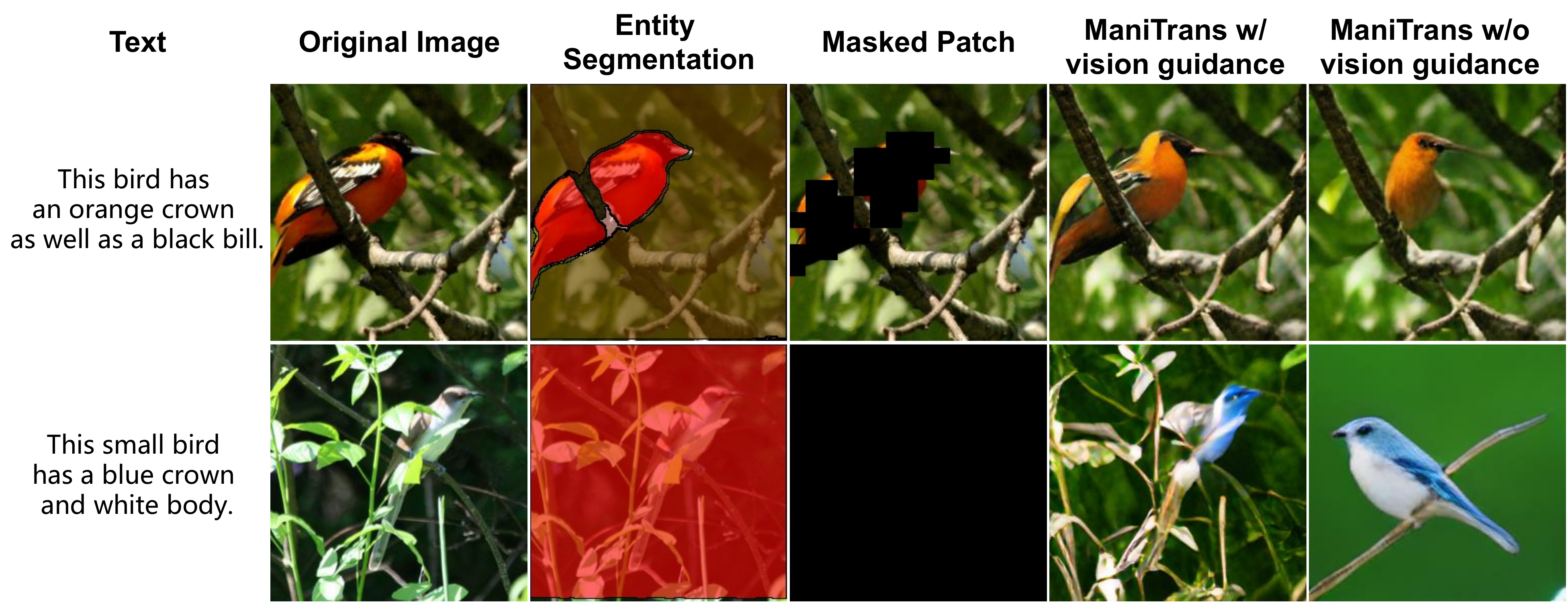}
% \vspace{-0.15in}
\caption{Our failed cases on CUB. The first row shows a failed case when without vision guidance.
With the entity shape information of vision guidance, our ManiTrans can manipulate the whole bird according to the text. While, without the entity shape information across the two-part segmentation, ManiTrans only generates a bird head within the limited right part region. The second row shows a failed case in preserving the background when without vision guidance. The bird with a white belly of the second row is difficult to be discriminated with the bright leaves, and the entity segmentation recognizes the whole image as an entity. Thus, without the vision guidance, our ManiTrans did a generation task and lost the original background. Most failed cases on CUB happen in these two situations.}
\label{fig:supp_failed_cub}
% \vspace{-0.15in}
\end{figure*}

\begin{figure*}
\centering
\includegraphics[width=\textwidth]{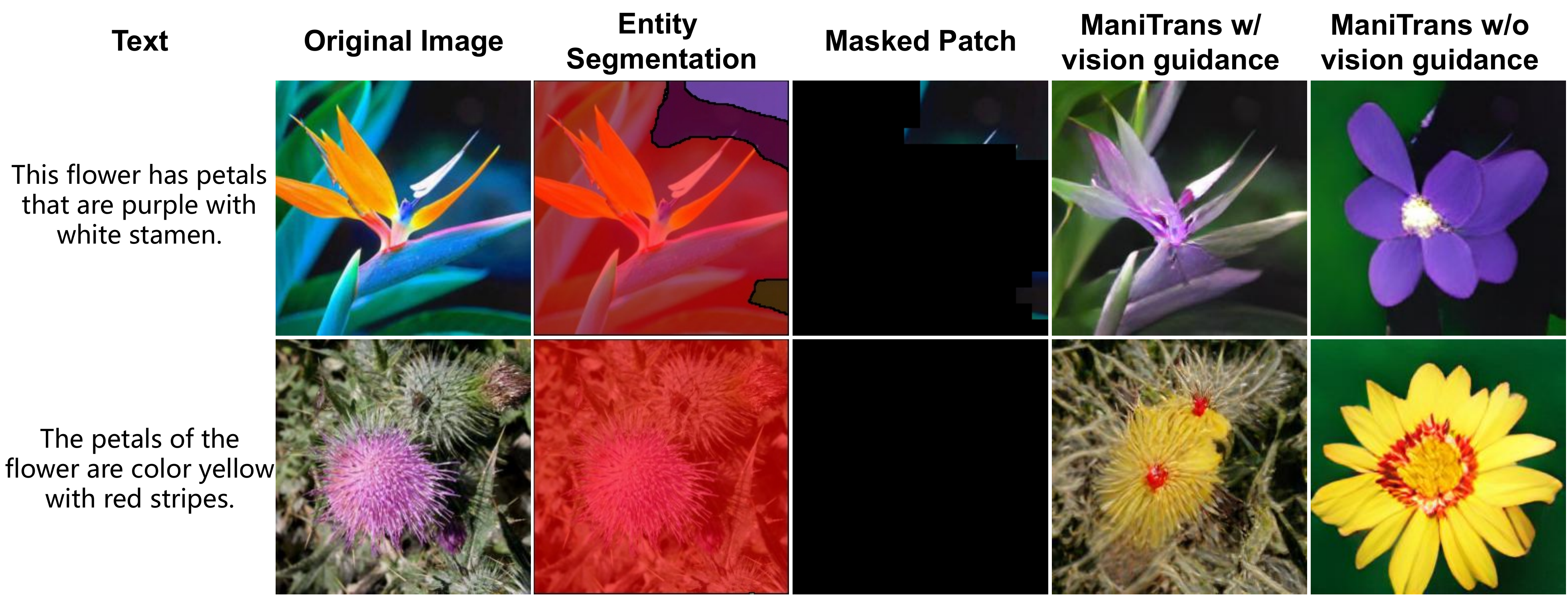}
% \vspace{-0.15in}
\caption{Our failed cases on Oxford. Most failed cases for flower manipulation fail in background preservation, for the flowers in Oxford are too large and difficult to discriminate with the background, especially the background of green leaves. Here two examples change the background and the flower at the same time, either with vision guidance or not.}
\label{fig:supp_failed_oxford}
\end{figure*}

\begin{figure*}
\centering
\includegraphics[width=\textwidth]{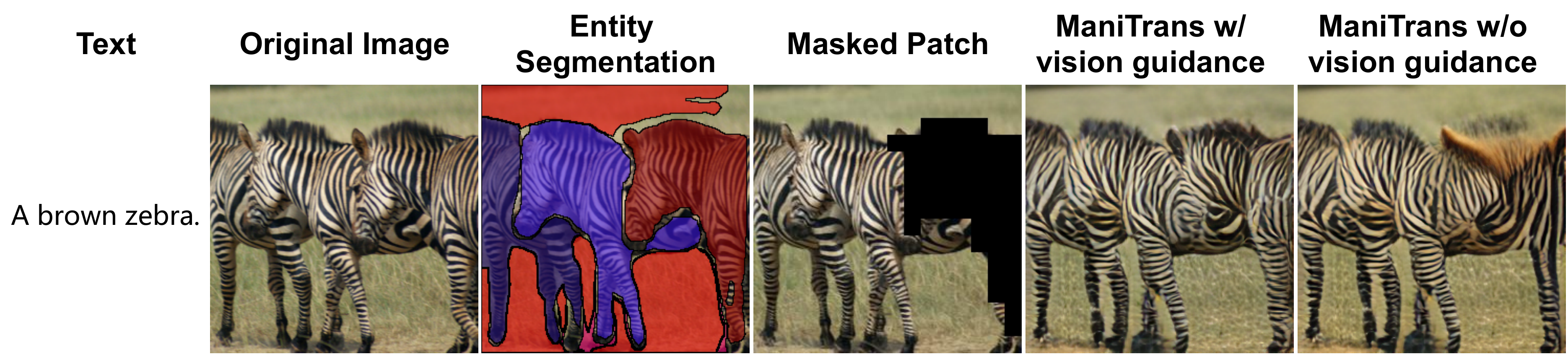}
% \vspace{-0.15in}
\caption{Our failed cases on COCO. Here we aim to manipulate one zebra on the image. However, confused by another preserved zebra next to the tokens to be manipulated, ManiTrans generates a brown body for the nearest zebra instead of generating a whole brown zebra. To manipulate an entity that is overlapped by other entities in the same category is to be improved.}
\label{fig:supp_failed_coco}
% \vspace{-0.15in}
\end{figure*}

We additionally present and discuss failed cases on the CUB (Fig.~\ref{fig:supp_failed_cub}), Oxford (Fig.~\ref{fig:supp_failed_oxford}) and COCO (Fig.~\ref{fig:supp_failed_coco}), where the black patches in the \textbf{Masked Patch} are the selected entity tokens by our semantic alignment module, to be manipulated, roughly reflected on the original image.

% \end{mytitlepage}

\end{document}